\documentclass[letterpaper, 10 pt, conference]{ieeeconf}  % Comment this line out
\IEEEoverridecommandlockouts                              % This command is only
\usepackage[utf8]{inputenc}
\usepackage[T1]{fontenc}
\usepackage{amsmath}
\usepackage{hyperref}
\title{\LARGE \bf
A Novel Weighted Ensemble Learning Based Agent for the Werewolf Game 
}

\author{ 
        \parbox{3 in}{ \centering Mohiuddeen Khan
        \\
        Department of Computer Engineering\\
         Aligarh Muslim University\\
             Aligarh, India\\
        {\tt\small mkhan42@myamu.ac.in}
        }
        \hspace*{ 0.5 in}
        \parbox{3 in}{\centering Claus Aranha\\
        
        Department of Computer Sciences\\
         University of Tsukuba\\
        Tsukuba, Japan\\
         {\tt\small caranha@cs.tsukuba.ac.jp}
        }
}

\usepackage{graphicx}
\usepackage{tabularx}
\usepackage{multirow}
\begin{document}

\maketitle
\thispagestyle{empty}
\pagestyle{empty}

%%%%%%%%%%%%%%%%%%%%%%%%%%%%%%%%%%%%%%%%%%%%%%%%%%%%%%%%%%%%%%%%%%%%%%%%%%%%%%%%
.\begin{abstract}

Werewolf is a popular party game throughout the world, and research on its significance has progressed in recent years. The Werewolf game is based on conversation, and in order to win, participants must use all of their cognitive abilities. This communication game requires the playing agents to be very sophisticated to win. In this research, we generated a sophisticated agent to play the Werewolf game using a complex weighted ensemble learning approach. This research work aimed to estimate what other agents/players think of us in the game. The agent was developed by aggregating strategies of different participants in the AI Wolf competition and thereby learning from them using machine learning. Moreover, the agent created was able to perform much better than other competitors using very basic strategies to show the approach's effectiveness in the Werewolf game. The machine learning technique used here is not restricted to the Werewolf game but may be extended to any game that requires communication and action depending on other participants

\end{abstract}

%%%%%%%%%%%%%%%%%%%%%%%%%%%%%%%%%%%%%%%%%%%%%%%%%%%%%%%%%%%%%%%%%%%%%%%%%%%%%%%%
\section{Introduction}

In complete information games like Chess and Shogi, artificial intelligence has already defeated top-level human players \cite{8816}. 
%
% Communication or communicative intelligence skills, which are widely used in board games and card games, have not been evaluated in comparison to earlier game-related obstacles. 
On the other hand, board games and card games that require communication or communicative intelligence skills are still an open research challenge.
People talk with other players while playing board and card games, and some games are referred to as "communication games" because they are advanced through conversation. 

%%What is AI WOlf???
"Werewolf" is one such conversation game that necessitates the use of our high intellectual abilities, including the capacity to assess an individual's history and judge their readiness to collaborate or be persuaded only via dialogue \cite{first_one}. AI Wolf is undeniably a game with a lot of research potential apart from artificial intelligence because it demands so many different fields of study. There are also various game competitions promoting and enhancing AI and communication in games, such as the Lemonade Stand Game Competition \cite{lemonade}.

% The AI werewolf game is based on a story as follows: “It's a village story,” says the narrator. Werewolves have come, capable of transforming into and devouring people. During the day, the werewolves seem as people, and at night, they attack the villagers one by one. Fear, apprehension, and scepticism about werewolves begin to develop. The villagers determine that anyone suspected of being werewolves shall be executed one by one"
The Werewolf game is based on the following story: "In one village, werewolves have come, capable of transforming into people and devouring them. During the day, the werewolves are seen as people, and they attack the villagers one by one at night. Fear, uncertainty, and doubt about the werewolves begin to develop. The villagers decide that anyone suspected of being a werewolf must be executed". In short, Werewolf is a social deduction game, where the villagers must discover who the werewolves are, and the werewolves must hide in the group.

In the Werewolf game, participants must use all of their cognitive abilities. In comparison to complete information games like Chess or Shogi, participants in the Werewolf game must conceal information. Each player gathers hidden knowledge from other players words and actions, then hides it to achieve their goals.  
%There is a lot of research going in the AI Wolf including \cite{second_one} \& \cite{third_one}.
There is a lot of research going about using AI to play the Werewolf game, which we call the "AI Wolf." Some examples of AI Wolf research include modeling players using statistics~\cite{second_one} and using BDI models to train AI agents in the werewolf game~\cite{third_one}. Additional research work includes estimating the player's role by using the number of interruptions and the length of utterances of the players \cite{utter} and using nonverbal information \cite{nonverbal}.

% Different agents made by different participants play the game.
In this work, we made an AI Wolf agent who can predict the hostility of the other agents towards him (i.e., our agent can predict which agent is going to vote for him in the game for elimination) using the talk patterns in the game. The result is of much greater significance as knowing who the hostile agents are can help design different strategies against them. The agent learns from the strategies of different agents in the game, using a novel model based on a weighted modified bootstrap aggregating ensemble, which will be described in detail later.
% The agent learns from the strategies of different agents in the game using a novel machine learning technique described in detail. The approach is not limited to AI Wolf, but to any other game where one player can learn from different players and combine their strategies/experiences to make a more sophisticated player/agent.

\section{Theory and Background}

\subsection{The Werewolf game}

The Werewolf game is a communication game that necessitates the use of various AI technologies, including multi-agent coordination, knowledge of the theory of mind, and much more \cite{first_one}. Each player has a role to perform in the game, and the goal is to win through communicating, misleading, and detecting lies in other players' messages. Players are divided into two teams in the Werewolf game: the werewolf team and the villager team. Every player is assigned a role respective to their team.\\ Those who play the role of werewolves know who the werewolves are, but players who play other roles don't. The game is divided between day and night periods. During the day phase, the participants conduct a lie-filled conversation and vote on who will be executed. During the night phase, the werewolves attend a secret meeting to choose one player to kill. 
% The execution occurs at the end of the day phase. The werewolves kill a player during the night period. 
Those who survive then move on to the next day phase. The villager's team has some players (ex. SEER, MEDIUM) with special abilities described later.\\
The victory condition for the villagers is to eliminate all the werewolves, whereas the werewolves team wins if the number of wolves is equal to or greater than the number of villagers. Fig.1 displays the flow of the Werewolf game. The following is the list of roles in the AI Wolf game:-\\

\begin{itemize}

\item Villager - This role belongs to the villagers team. This role has no special ability.
\item Werewolf - This role belongs to the werewolves team. The werewolf role has the power to attack any agent from the villager team at the night. The werewolf can discuss among other werewolves who to attack in the night.
\item Seer - This role belongs to the villagers team, This role has a special ability to find the role of any player in the game by inspecting him at night.
\item Medium - This role belongs to the villagers team. This role has a special ability to ascertain whether the player who died the day before by voting was a werewolf or not.
\item Bodyguard - This role belongs to the villagers team. This role has a special ability to guard an agent from the attack of the werewolves in the night. The guarded agent then, won't be eliminated by the attack of the werewolves.
\item Possessed - This role belongs to the werewolves team. The possessed has no special ability and they try to play in a way such that the werewolves win.

\end{itemize}

\begin{figure}
\centering
\includegraphics[width=8cm,height=14cm]{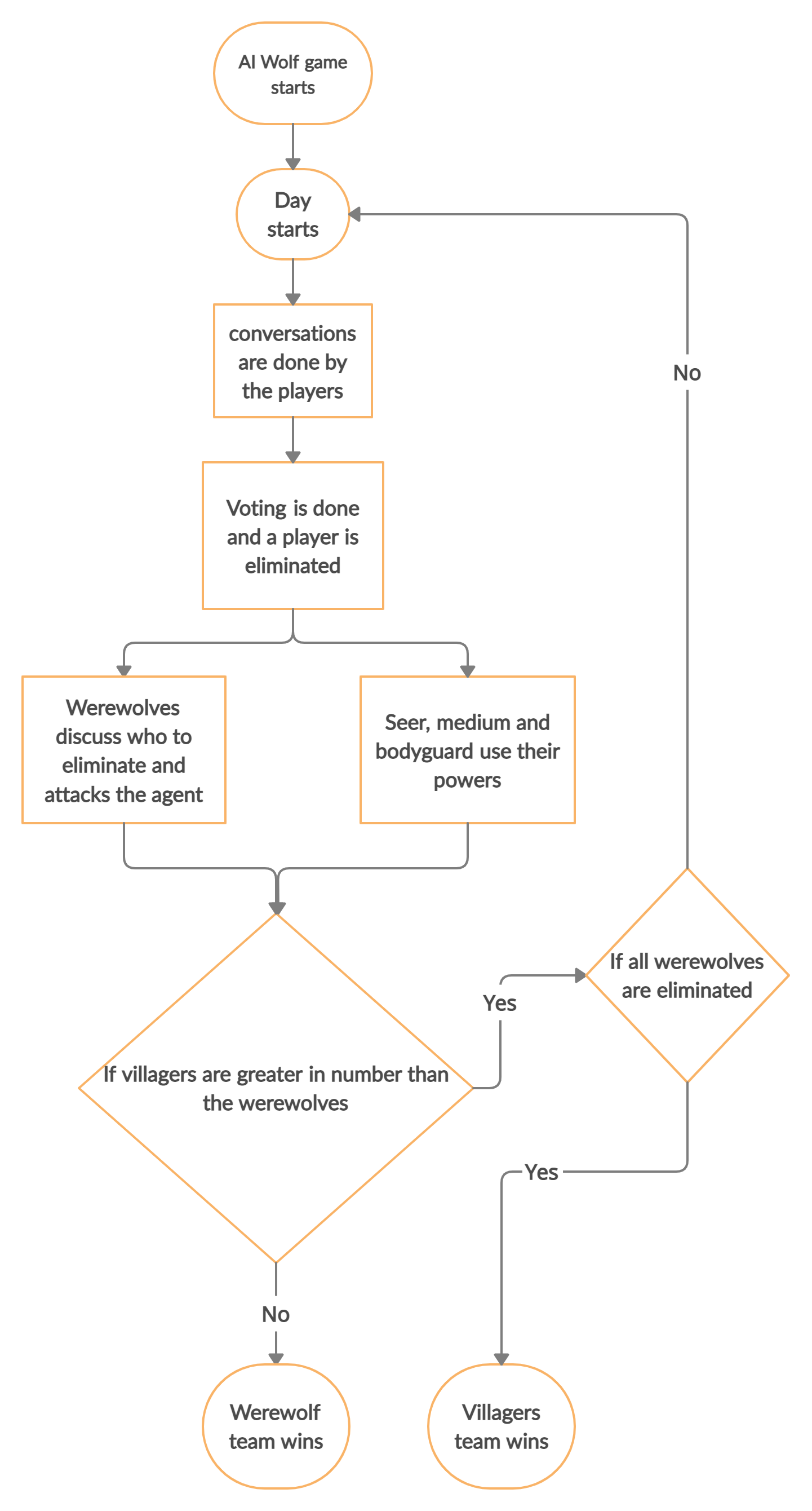}
\caption{Werewolf game flow }

\end{figure}

% \begin{table}[ht]
% \caption{Roles and abilities in AI Wolf} % title of Table
% \centering % used for centering table
% \begin{tabular}{c c c c} % centered columns (4 columns)
% \hline\hline %inserts double horizontal lines
% Role & Team & Ability  \\ [0.5ex] % inserts table
% %heading
% \hline % inserts single horizontal line
% Villager &  Townsfolk team  & No special ability \\ % inserting body of the table

% Werewolf & Werewolf team & \\ 
% Seer & & \\
% Medium & & \\
% Bodyguard & & \\
% Possessed & & \\

% \hline %inserts single line
% \end{tabular}
% \label{table:nonlin} % is used to refer this table in the text
% \end{table}

\subsection{AI Wolf agents and competition}

The AI wolf competition is held every year in which different agents participate representing different teams \cite{first_one}. The game is played in 2 versions, viz. 5 player and 15 player games.  The participating agents were essential to this research work as we used the agents from the past competition for our research. Our agent will try to learn from different agents and will try to combine their strategies to become a robust and sophisticated agent itself. The agents used in this research work were: Takeda, Wasabi, Viking, and Daisyo. We used these agents in our experiment because they excelled in the AI Wolf competition and would provide better statistics in the experiment. An agent named "Sample" was also used, which is an agent available on the AI Wolf website.

\subsection{Ensemble learning} Ensemble Learning is an approach that seeks to give better predictive accuracy by combining numerous machine learning models constructed using various learning algorithms \cite{fourth_one}. To get a better prediction, it simply averages the predictions of multiple models using different ways. There are mainly two types of ensemble methods, i.e., bagging and boosting \cite{boost}. Our research work includes bootstrap aggregating (bagging), which is described in detail later. The basic idea of ensemble learning in games is to use the results from different strategies or combine predictions from different models.

\subsection{Bootstrap aggregating (bagging)}

Bootstrap aggregation, also known as bagging, is an ensemble learning method that varies the training data to find a varied collection of ensemble members \cite{fifth_one}. Both regression and statistical classification can benefit from the bagging strategy. Bagging is used with decision trees to significantly improve model stability by enhancing accuracy and reducing variance, removing the problem of overfitting.

As shown in Fig.2, bagging works by generating a certain number of models and corresponding datasets from the primary dataset by sampling and replacement. The models are then trained on the corresponding datasets. After the models have been trained, they are given a sample to test, and each model produces an output value. After that, the output values from several models are merged using hard/soft voting to produce a final outcome. Because it mixes the outputs from various models, bagging produces better results on classification problems than traditional algorithms.

\begin{figure}
\centering
\includegraphics[width=8cm,height=9cm]{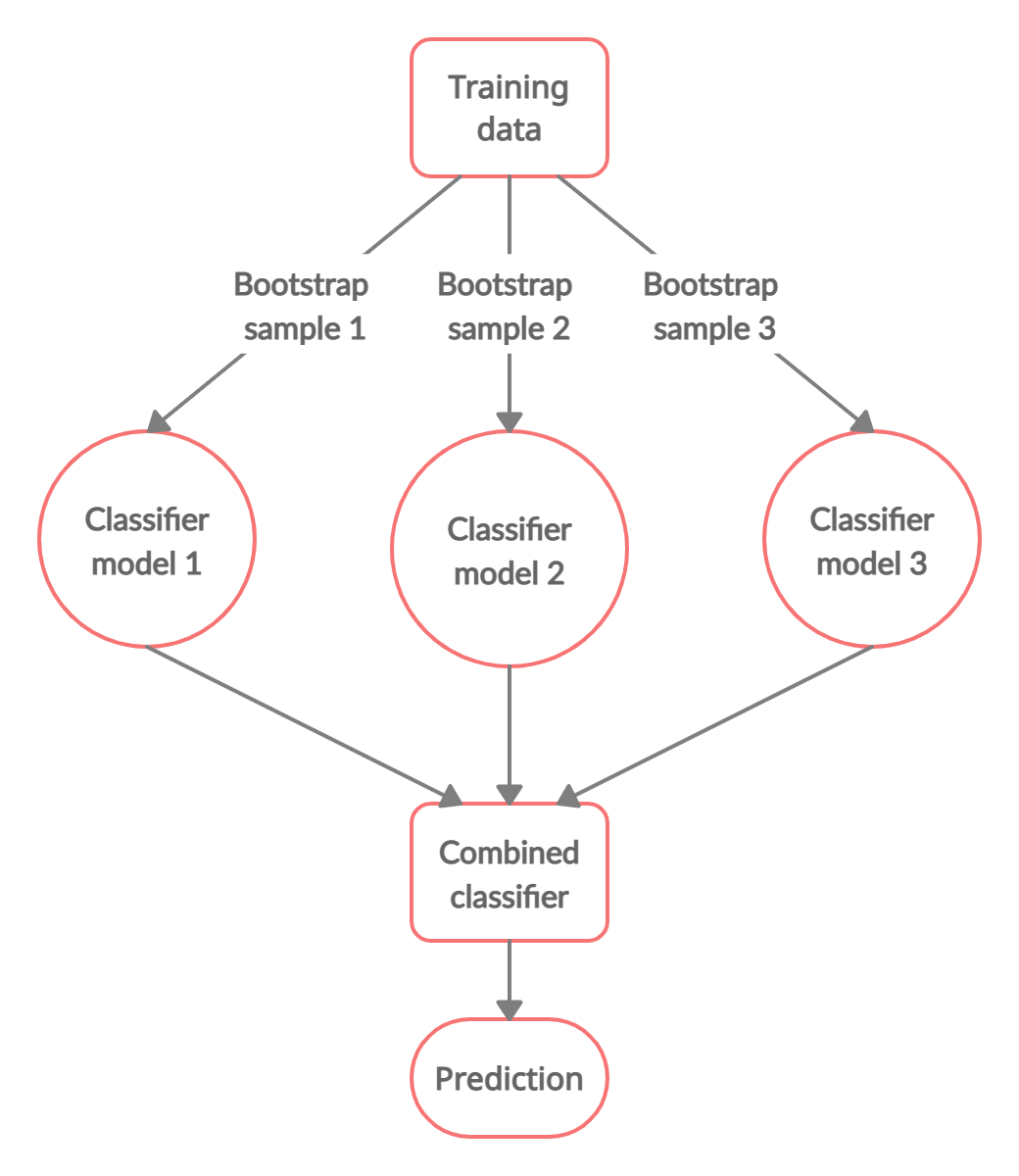}
\caption{Bootstrap aggregating (bagging)}

\end{figure}

\subsection{Weighted ensemble voting} 

Ensemble learning in binary classification problems combines the results obtained from different models by averaging the outputs of different models either by hard or soft voting. 
The projected output class in hard voting is the one with the most votes, i.e., the one with the highest likelihood of being predicted by each classifier. For example, if three classifiers give outputs for a binary classification problem as [0,1,1], then the final result will be 1 as it is predicted by the majority of classifiers i.e., 2.

The output class in soft voting is the prediction based on the average probability assigned to that class. Assume that given some input to three models, the prediction probability for class 1 = (.90, .80, .70) and 0 = (.10, .20, .30). So, with an average of 0.80 for class 1 and 0.20 for class 0, class 1 is clearly the winner because it had the highest probability averaged by each classifier.

The problem with hard and soft voting is that if a model is very good and robust compared to other models in the ensemble, then its vote is counted equally to all the models in the ensemble. It is wise to provide a hierarchical system to the models so that each model has its own weight and a larger contribution corresponding to its robustness. We used and modified an approach given in \cite{8907028} to create our agent. Equation (1) gives the formula of assigning weights to the models based on their robustness. The approach is shown in Fig.3 and it works as follows:\\
Initially, all the weights are equal to 1. All the instances in the validation set are traversed, and the weights for the classifiers are updated using the Equation (1). For example, refer to Tables I and II. An example dataset is given in Table II to update the weights of the three classifiers in Table I. For the first instance, classifier 3 predicts the correct output as verified from Table II, whereas classifiers 1 and 2 give a wrong output class. The weights of the first two classifiers are not changed due to penalty, whereas a reward is given to the third classifier by adding a value given in Equation (2). A similar process is done for three instances, and the final weights are calculated as shown in Table I.  \\

\begin{equation}
    w_{i,j} = \left\{ \begin{array}{c c c} w_{i-1,j} + r & \mbox{for correct prediction} & 
    
    \\ w_{i-1,j} & \mbox{for incorrect prediction} \end{array} \right.
\end{equation}

\begin{equation}
    r = Y_{wrong}/n \\
\end{equation}

$where:$ \\
$w = weight, r = reward$\\
$Y_{wrong}$ $=$ $number$ $of$ $classifiers$ $that$ $did$ $wrong$ $prediction$\\
$n$ $=$ $number$ $of$ $classfiers$ \\

% \begin{table}[h]
% \caption{Metrics table}
% \label{table_example}
% \begin{center}
% \begin{tabular}{|c||c||c||c|}
% \hline
%  Instances/Classifiers & Classifier 1 & Classifier 2 & Classifier 3 \\
% \hline
% Initial weights & 1 & 1 & 1  \\
% \hline
% \hline
% 1 & 1 + 0/3 = 1 & 1 + 0/3 = 0 & 1 + 2/3 = 1.67  \\
% \hline
% \hline
% 2 & 1 + 0/3 = 1 & 1 + 1/3 = 1.33 & 1.67 + 1/3 = 2 \\
% \hline
% \hline
% 3 & 1 + 0/3 = 1 & 1.33 + 0/3 = 0 & 2 + 2/3 = 2.67 \\
% \hline

% \hline
% Final weights & .440 & .460 & .620  \\
% \hline

% \hline
% \end{tabular}
% \end{center}
% \end{table}

\begin{table}[h]
\caption{changes in weight through each instance. Adapted from \cite{8907028}}
\label{table_example}
\begin{center}
\begin{tabular}{ |p{2cm}|p{1.5cm}|p{1.5cm}|p{1.5cm}|p{1.5cm}|  }
%  \hline
%  \multicolumn{4}{|c|}{Country List} \\
 \hline
 Instances/ Classifiers& C1 prediction & C2 prediction & C3 prediction \\
 \hline
   1   & 1 + 0 = 1  & 1 + 0 = 1&  1 + 2/3 = 1.67\\
 \hline
 2  &   1 + 0 = 1  & 1 + 1/3 = 1.33   & 1.67 + 1/3 = 2\\
 \hline
 3 &  1 + 0 = 1 & 1.33 + 0 = 1.33  &  2 + 2/3 = 2.67\\
 \hline
 Final weights & 1 & 1.33 &  2.67\\
 \hline
\end{tabular}
\end{center}
\end{table}

\begin{table}[h]
\caption{EXAMPLE DATASET}
\label{table_example}
\begin{center}

\begin{tabular}{ |p{2cm}|p{1cm}|p{1cm}|p{1cm}|p{1cm}|  }
%  \hline
%  \multicolumn{5}{|c|}{Country List} \\
 \hline
 Instances/ Classifiers& C1 prediction & C2 prediction & C3 prediction & Actual Class\\
 \hline
 1   & X   & X &   Y & Y\\
 \hline
 2 &   Y  & X   & X & X\\
 \hline
 3 & Y & Y &  X & X\\

 \hline
\end{tabular}
\end{center}
\end{table}

After the weights were assigned to every model by going through every instance of the validation set, the ensemble model was fed with test data, with every instance of test data going into every model. Every model predicts an output that is multiplied by its corresponding weight. For example, referring to the previous two Tables, if we have 3 models in the ensemble and the weights calculated for the 3 models are [1,1.33,2.67] and the outputs from different classifiers are [1,1,0] then the total output for class 1= 1*1 + 1.33*1= 2.33 and total output for class 0 = 2.67*1 = 2.67. As 2.67>2.33 so, the final output will be class 0. Had this been simple hard voting, we would've got class 1 as the output due to the majority of classifiers giving output as class 1.\\
The advantage of the discussed approach is that it doesn't give an equal contribution to a poor and good model. Every model in the ensemble will contribute to the output based on its performance.

\begin{figure}
\centering
\includegraphics[width=8cm,height=9cm]{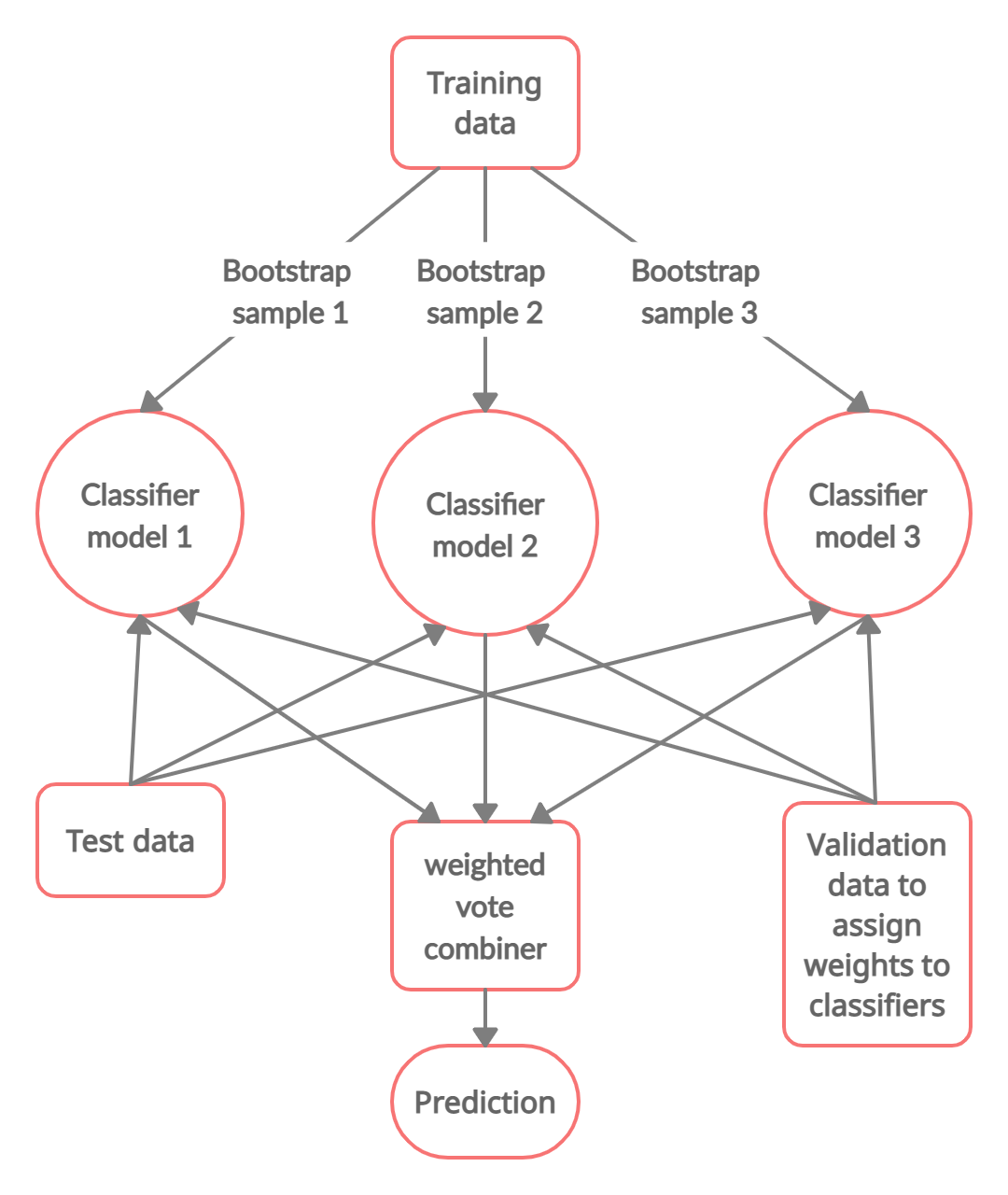}
\caption{Weighted ensemble approach by \cite{8907028} }

\end{figure}

\section{Proposed Method}

We propose a novel weighted modified bootstrap aggregating ensemble model to make an agent in this research. 
The proposed agent can combine the experiences and strategies of different agents from the AI Wolf competition and can anticipate which agent in the competition will vote for him at the end of the day in the game. An implementation of the proposed method is available at
% \href{https://github.com/mohikhan/Weighted-ensemble-voting-approach-for-AI-Wolf}{github part 1}
\href{https://github.com/}{github part 1 (link)}.
% and
% % \href{https://github.com/mohikhan/Hostility-prediction-in-AI-Wolf}{github part 2}
% \href{https://github.com/}{github part 2 (link)}

Knowing who will vote for our agent at the end of the day in the game can be very beneficial for us as after predicting the agent who will vote for us, we can try to convince that agent to not vote for us or we can convince other agents to eliminate that hostile agent from the game, etc.
\subsection{Data Attributes}

To estimate which agent is going to vote for us, we are going to predict a binary value for every agent in the game if whether that agent is going to vote for us or not. We are going to record the positive sentences, negative sentences, and the length of the negative sentences every day for every agent. The positive sentences are the number of sentences that are considered friendly that an agent speaks about us. The program estimates if a sentence is positive or negative by finding patterns and keywords. An example of a positive sentence is:\\

"ESTIMATE Agent[{myid}] VILLAGER"\\

which means that the agent is saying that he thinks I am a villager. Note how the keywords "ESTIMATE" followed by the keyword "VILLAGER" is classifying the sentence in a positive category.

The negative sentences are the number of sentences that are considered hostile that an agent speaks about us in a complete day. Some examples of negative sentences are:\\

"ESTIMATE Agent[{myid}] WEREWOLF"\\

which means that some agent is saying that he estimates me as a werewolf. Here we can note how the keywords "ESTIMATE" followed by the keyword "WEREWOLF" is classifying the sentence into a negative category. Another example could be:\\

"REQUEST ANY (VOTE Agent[{myid}])"\\
 
which means that some agent is requesting any agent to vote for me.

The negative length is the sum of the length(s) of the negative sentences(s) that the agent speaks about us. For example, The negative length for the sentence\\

"ESTIMATE Agent[{myid}] WEREWOLF" \\

will be 29 as this string contains 29 characters including spaces.

The value for positive sentences and negative sentences can also be zero, which can happen when no player spoke about us in the whole day or if our agent has died.  

After collecting these three values for every agent, we collected a binary value (0/1) called "Vote(Yes/No)," which signified if that particular agent voted for us or not.

\subsection{Data Collection} 

% For the proposed model, one instance of the input data corresponds to one game-day of talks, the three parameters extracted from these talks, and the vote result, as described in table~\ref{table:nonlin}
% To collect this data, we executed 500 games of one "target" agent against four basic agents, who just spoke simple sentences.
% We repeated this process for every agent that participated in the 2nd international competition. For each agent, we produced one csv file with parameters, and whether that agent was voted for.

%Change the order of the para
To collect the data from every participating agent of the competition, we generated one CSV file for every agent by targeting that agent for 500 games and storing those values in the CSV file. By targeting, we mean that we made a very basic agent who just spoke simple sentences and collected the values of the three parameters by continuously tracking the targeted agents (every agent from the AI Wolf competition). Every single instance signifies one day in the game where the end result is whether the targeted agent voted for us or not at the end of the day.

Some examples of instances of data collected on targeting the "Takeda" agent is shown in table III.

\begin{table}[ht]
\caption{Dataset generated by targeting Takeda} % title of Table
\centering % used for centering table
\begin{tabular}{p{1cm} p{1cm} p{1cm} p{1cm} p{1cm}} % centered columns (4 columns)
\hline %inserts double horizontal lines
Game/Day & Negative talks & Positive talks & Negative length & Vote  (Yes/No)  \\ [0.5ex] % inserts table
%heading
\hline % inserts single horizontal line
 Game1 Day1 & 2 &  0 & 23 & 1   \\ % inserting body of the table
\\
 Game1 Day2 & 0 & 2 & 0 & 0 \\
\\
Game2 Day1 & 1 & 0 &  8 & 1 \\
\\
.... &.... & .... & ... & ...\\
\\
so on & so on & so on & so on & so on\\
\hline %inserts single line
\end{tabular}
\label{table:nonlin} % is used to refer this table in the text
\end{table}

The first instance shows a training instance in which the Takeda agent said two negative sentences with a total length of 23 characters in that particular day discussion. At the end of the day, the Takeda agent voted for our agent as the Vote value is 1. The third training instance is almost the same as the first training instance except for the fact that this time Takeda said only one negative sentence with a total length of 8 characters, and at the end of the day, he voted for us. The second training instance, however, implies that Takeda said one positive sentence about us, and at the end of the day, it didn't vote for our agent, which is pretty clear as if someone speaks positive about us, it is unlikely that that agent will vote for us to get eliminated.

\subsection{Ensemble training}
We trained five separate random forests on the five agents (Takeda, Wasabi, Viking, Daisyo, and Sample) from the AI Wolf competition, respectively, as our ensemble models. The fact that the ensemble model contains full random forests inside it is a crucial difference from conventional ensemble learning, which uses decision trees. The data collected from the five agents was split into 70\% training and 30\% validation. The 70\% data was used to train every model separately with their training data. For example, The Takeda random forest model was trained on 70\% training data collected from the Takeda agent. Similarly, all other models were trained on their respective 70\% data. After the training, the 30\% data from all the five agents was collected at one place and sampled. This 30\% data collected from the five agents was used for updating the weights of the models using Equation (1) and acted as validation data for them.

\subsection{Out of Sample testing and agent validation}
To test the model metrics, data was collected on a participant agent named "Tomato" and was used as a test set to evaluate the ensemble model. After calculating the metrics, the agent was deployed in the real AI wolf game with a very basic strategy to test the ensemble model to see how it was performing.

\begin{figure}
\centering
\includegraphics[width=8cm,height=11cm]{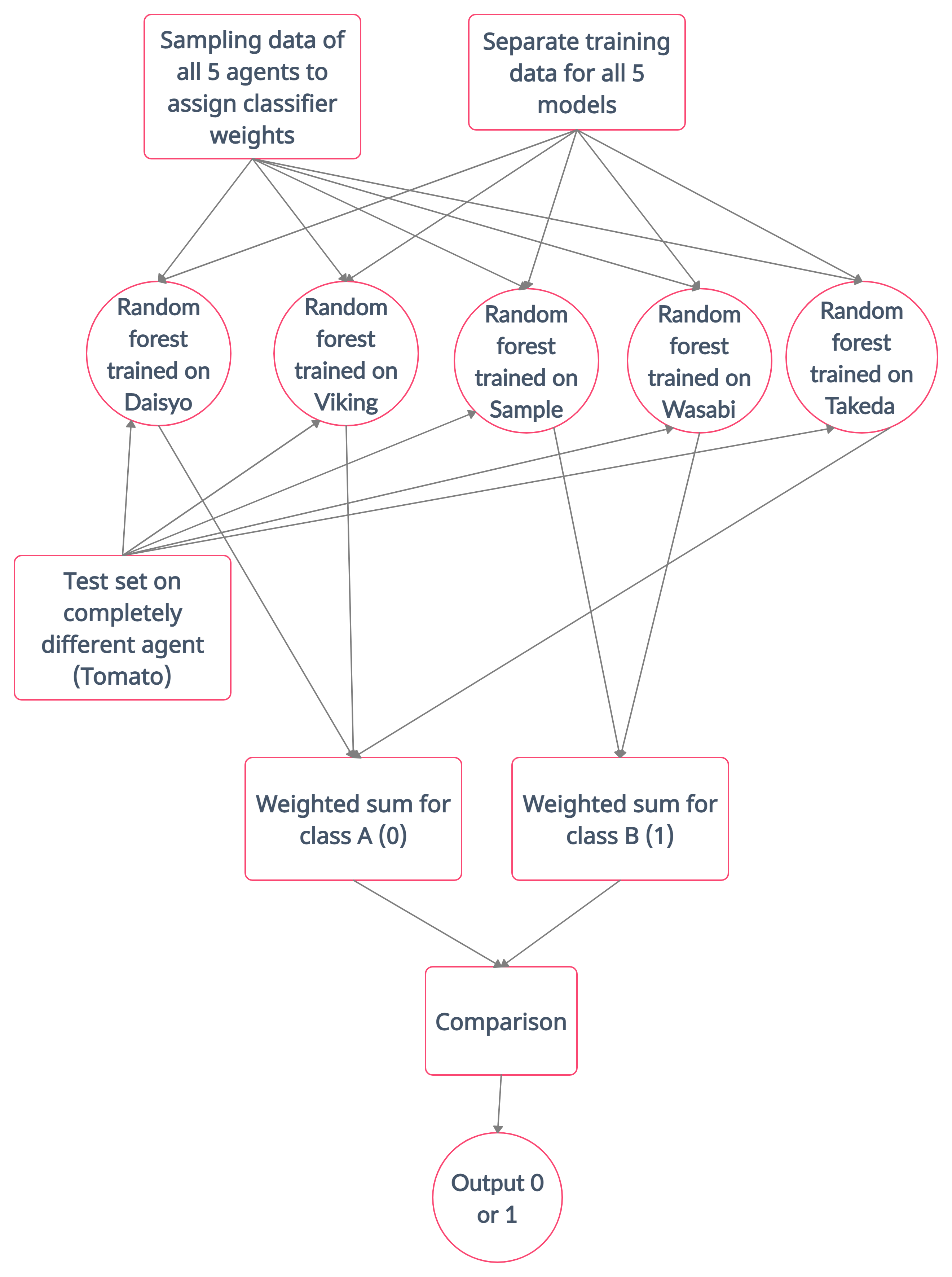}
\caption{Proposed ensemble model}
\end{figure}

\section{Experiment}
The first experiment was initiated to see the metrics of the ensemble model on the data from the agent, which the ensemble model has not seen previously. Described in the above section, the test set was used from "Tomato" and was used to see the metrics. After using the validation set sampled from the data of 5 agents, the final weights of the model are shown in Table IV.
%add classifier weights term here

\begin{table}[ht]
\caption{Final weights generated} % title of Table
\centering % used for centering table
\begin{tabular}{c c } % centered columns (4 columns)
\hline\hline %inserts double horizontal lines
Model name & Classifier weight  \\ [0.5ex] % inserts table
%heading
\hline % inserts single horizontal line
Takeda random forest &  25.39   \\ % inserting body of the table

Wasabi random forest & 22.20 \\ 
Sample random forest & 24.79 \\ 
Viking random forest & 26.39 \\ 
Daisyo random forest & 15.39 \\ 

\hline %inserts single line
\end{tabular}
\label{table:nonlin} % is used to refer this table in the text
\end{table}

The ensemble model with five random forests was deployed on the test set from the "Tomato" agent to predict the binary value 0 (not voted) or 1 (voted for us). Notice that for every test instance from "Tomato," all the five models predicted a binary value. The binary value was then multiplied by the weights of the corresponding model given in Table IV, followed by hard voting to calculate the output. The proposed approach is also showed in Fig.4. 

% For example consider the following test instance from "Tomato" dataset:- \\

% \begin{table}[ht]
% % \caption{Dataset generated by targeting Takeda} % title of Table
% \centering % used for centering table
% \begin{tabular}{c c c c} % centered columns (4 columns)
% \hline\hline %inserts double horizontal lines
% Negative talks & Positive talks & Negative length & Vote(Yes/No)  \\ [0.5ex] % inserts table
% %heading
% \hline % inserts single horizontal line
% 2 &  23 & 0 & 1   \\ % inserting body of the table
% \hline %inserts single line
% \end{tabular}
% \label{table:nonlin} % is used to refer this table in the text
% \end{table}

% Suppose when the above instance was fed into the ensemble model then every model will give one output for ex: [1,0,0,1,1] were the output values given from every corresponding model in the ensemble.

We also checked whether the accuracy is changing with the number of trees or not, and the model was then tried for a different amount of trees, and the results are shown in Table V. The results clearly show that the model's accuracy did not change significantly with the number of trees used. Therefore, we used 10 trees in each random forest to save excessive computation.

\begin{table}[h]
\caption{Metrics table}
\label{table_example}
\begin{center}
\begin{tabular}{|c||c||c|}
\hline
No of trees & Accuracy(\%) & MAE \\
\hline
10 & 80.34 & .1965 \\
\hline
\hline
20 & 79.48 & .2051 \\
\hline
\hline
30 & 82.05 & .1794 \\
\hline
\hline
40 & 80.34 & .1965\\
\hline
\hline
50 & 79.48 & .2051 \\
\hline
\hline
60 & 79.48 & .2051 \\
\hline
\hline
70 & 80.34 & .1965 \\
\hline
\hline
80 & 80.34 & .1965 \\
\hline
\hline
90 & 81.19 & .188 \\
\hline
\hline
100 & 80.34 & .1965 \\
\hline
\end{tabular}
\end{center}
\end{table}

After we were able to see that the ensemble model got good metrics on an agent that it had never seen before, we were ready to integrate the ensemble function into a real agent and play it in the AI Wolf game to see how it actually performed against other agents. We created a very basic agent which kept an eye on every agent and collected the data, i.e., positive sentences, negative sentences, and the length of the negative sentences for every agent everyday, and called the predict\_vote function, which calls the ensemble method at the end of the day to know who is going to predict for him. After knowing the agent(s) who are going to vote for him, we followed a simple strategy:

"Try to eliminate any agent who is going to vote for us as stated by the ensemble function. We also calculated a hate score which is nothing but a score that gets increased every time someone says negative things about us (calculated through the count of negative sentences). The hate score is useful as there might be times when there will be no agent to vote for us, and at that time, we would need somebody to vote for, so we will vote for the agent having the maximum hate score. Also, in case the ensemble model predicts two or more agents who are going to vote for us, or in case there are two or more agents whose hate scores are the same, we simply apply the hate strategy to a random agent among the two or more agents that are predicted by the ensemble/having same hate scores."

The same is shown in Fig.5 with our hate strategy "x" just defined above.
The reason for not programming a very complex agent was to see how the sole use of the ensemble model can help us in winning and also how useful the information is to predict which agent is going to vote for us.

\begin{figure}
\centering
\includegraphics[width=7cm,height=6cm]{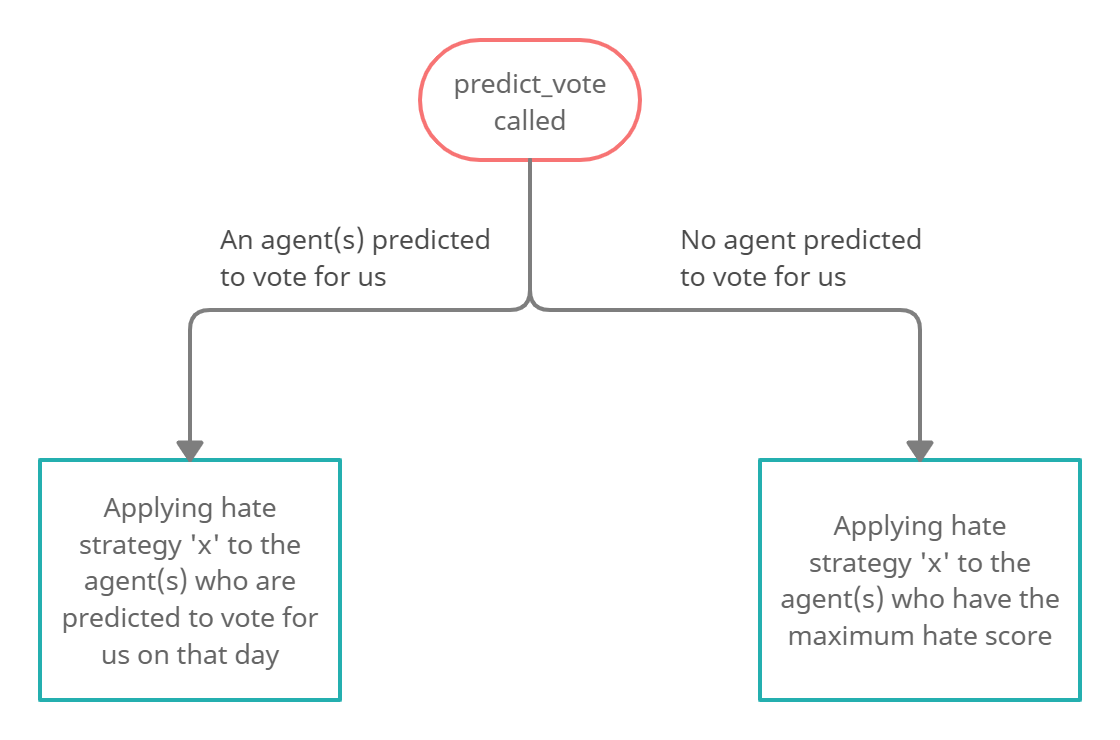}
\caption{Ensemble agent strategy }

\end{figure}

\begin{table}[h]
\caption{Metrics table for 5 player game}
\label{table_example}
\begin{center}
\begin{tabular}{|c||c||c||c||c|}
\hline
Agent & Iteration 1 & Iteration 2 & Iteration 3 & Average \\
\hline
Ensemble agent & .480 & .450 & .430 & .453  \\
\hline
\hline
Takeda & .640 & .630 & .690 & .653 \\
\hline
\hline
Wasabi & .470 & .460 & .420 & .450 \\
\hline
\hline
Tomato & .590 & .620 & .520 & .576\\
\hline
\hline
Daisyo & .440 & .490 & .490 & .473 \\
\hline

\hline
\end{tabular}
\end{center}
\end{table}

\begin{table}[h]
\caption{Metrics table for 15 player game}
\label{table_example}
\begin{center}
\begin{tabular}{|c||c||c||c||c|}
\hline
Agent & Iteration 1 & Iteration 2 & Iteration 3 & Average \\
\hline
Ensemble agent & .520 & .500 & .500 & .506  \\
\hline
\hline
Takeda & .500 & .600 & .480 & .526 \\
\hline
\hline
Wasabi & .320 & .460 & .380 & .386 \\
\hline
\hline
Tomato & .300 & .400 & .400 & .366\\
\hline
\hline
Daisyo & .440 & .460 & .620 & .506 \\
\hline

\hline
\end{tabular}
\end{center}
\end{table}

\section{Analysis}

Concluding the results of this research, an ensemble agent was deployed using a simple approach as stated in the previous section. As discussed earlier, the agent played the two types of AI Wolf competition games, viz. five-player, and fifteen-player games. For the five-player game, the agent played 100 games for three iterations, and the results are shown in Table VI. For the fifteen-player game, the agent played 50 games for three iterations, and the results are shown in Table VII. For every iteration, there is a corresponding value of the average winning rate. For example, In the 15 player game, iteration 1 has the average winning rate from 50 games of a particular player. The final average winning rate is the average of the three iterations for which the player is played. 

In comparison to the five-player game, the ensemble agent performed well in the fifteen-player game, as shown in Tables VI and VII. The agent in the fifteen-player game projected a significantly more accurate value than in the five-player game because the 15 player game has a much longer discussion than the five-player game owing to the larger number of participants. As a result of the longer conversations, our ensemble agent can acquire more data in a fifteen-player game (which is a test instance for the ensemble model) and forecast significantly more sophisticated output regarding whether a particular agent will vote for him or not.

\section{Conclusion}

In this work, we used a weighted ensemble learning approach to create an agent in the Werewolf game to compete with other players using the experiences of other competitors. The ensemble agent outperformed other agents in a 15-player game with very simple tactics, demonstrating the technique's effectiveness. The information needed to determine which agent will vote in our favor is critical. Despite a simple approach of targeting the agent who is expected to vote for us, we successfully defeated the sophisticated agents. Using our technique, a more complex strategy in the Werewolf game will definitely outperform other players. The projected knowledge will be used to construct a sophisticated agent in the future.

This method may be applied to any game in which the player can learn from the techniques of other players, not only the AI wolf game. In addition to games, employing a weighted method in ensemble learning may be quite beneficial.

\addtolength{\textheight}{-12cm}   % This command serves to balance the column lengths

%%%%%%%%%%%%%%%%%%%%%%%%%%%%%%%%%%%%%%%%%%%%%%%%%%%%%%%%%%%%%%%%%%%%%%%%%%%%%%%%

%%%%%%%%%%%%%%%%%%%%%%%%%%%%%%%%%%%%%%%%%%%%%%%%%%%%%%%%%%%%%%%%%%%%%%%%%%%%%%%%

%%%%%%%%%%%%%%%%%%%%%%%%%%%%%%%%%%%%%%%%%%%%%%%%%%%%%%%%%%%%%%%%%%%%%%%%%%%%%%%%

\bibliographystyle{plain}
\bibliography{bibliography.bib}

% \begin{thebibliography}{99} 

% \bibitem{c1} G. O. Young, ``Synthetic structure of industrial plastics (Book style with paper title and editor),'' 	in Plastics, 2nd ed. vol. 3, J. Peters, Ed.  New York: McGraw-Hill, 1964, pp. 15--64.
% \bibitem{c2} W.-K. Chen, Linear Networks and Systems (Book style).	Belmont, CA: Wadsworth, 1993, pp. 123--135.

% \end{thebibliography}

\end{document}